\documentclass[10pt,twocolumn,letterpaper]{article}

\usepackage[pagenumbers]{cvpr}      
\definecolor{cvprblue}{rgb}{0.21,0.49,0.74}
\usepackage[pagebackref,breaklinks,colorlinks,allcolors=cvprblue]{hyperref}
\usepackage{multirow}
\usepackage{comment}
\usepackage[table]{xcolor}

\def\confName{CVPR}
\def\confYear{2027}

\title{Counting Symbols on Technical Drawings: A Benchmark and Verified Rewards for MLLMs}

\author{Junxian Li$^{1}$\thanks{These authors contributed equally.}, Ruixuan Yang$^{2*}$, Zhenhua Xu$^{3}$ \\
\small{$^1$Shanghai Jiao Tong University, $^2$Xi'an Jiaotong University, $^3$Zhejiang University} \\
\texttt{\small{Corresponding to: lijunxian0531@sjtu.edu.cn}} 
}

\begin{document}
\maketitle
\begin{abstract}
    Today's MLLMs can count objects in natural images. However, when asking them how many bolt holes appear in a mechanical drawing or how many nitrogen atoms are in a chemical structure, they often struggle. Counting in technical drawings remains challenging, as objects can be small, densely packed, or even implied by drawing conventions. To study this problem, we introduce CountDrawBench, a benchmark including floor plans, electrical schematics, mechanical drawings, and molecular structures, with 1,574 counting questions and instance-level annotations where available. Besides evaluation, we introduce Code-Verified RL, a reinforcement learning framework that teaches MLLMs to count with executable feedback. Our method combines three complementary rewards: a count reward that checks the final answer, an instance-matching reward that verifies object locations, and a zoom-based reward that checks counting consistency across local image regions. All three are integrated into a unified GRPO objective. Extensive experiments reveal that even leading MLLMs struggle on CountDrawBench, with the best evaluated model achieving only 59.3\% accuracy. Our method improves Qwen3.5-9B from 49.3\% to 54.3\% accuracy, outperforming all evaluated open-source baselines. These results establish CountDrawBench as a challenging testbed for technical drawing understanding and demonstrate the value of code-verified rewards in these tasks. \underline{Code and data will be released upon acceptance.}
\end{abstract}

\section{Introduction}

\begin{figure}[t]
    \centering
    \includegraphics[width=\linewidth]{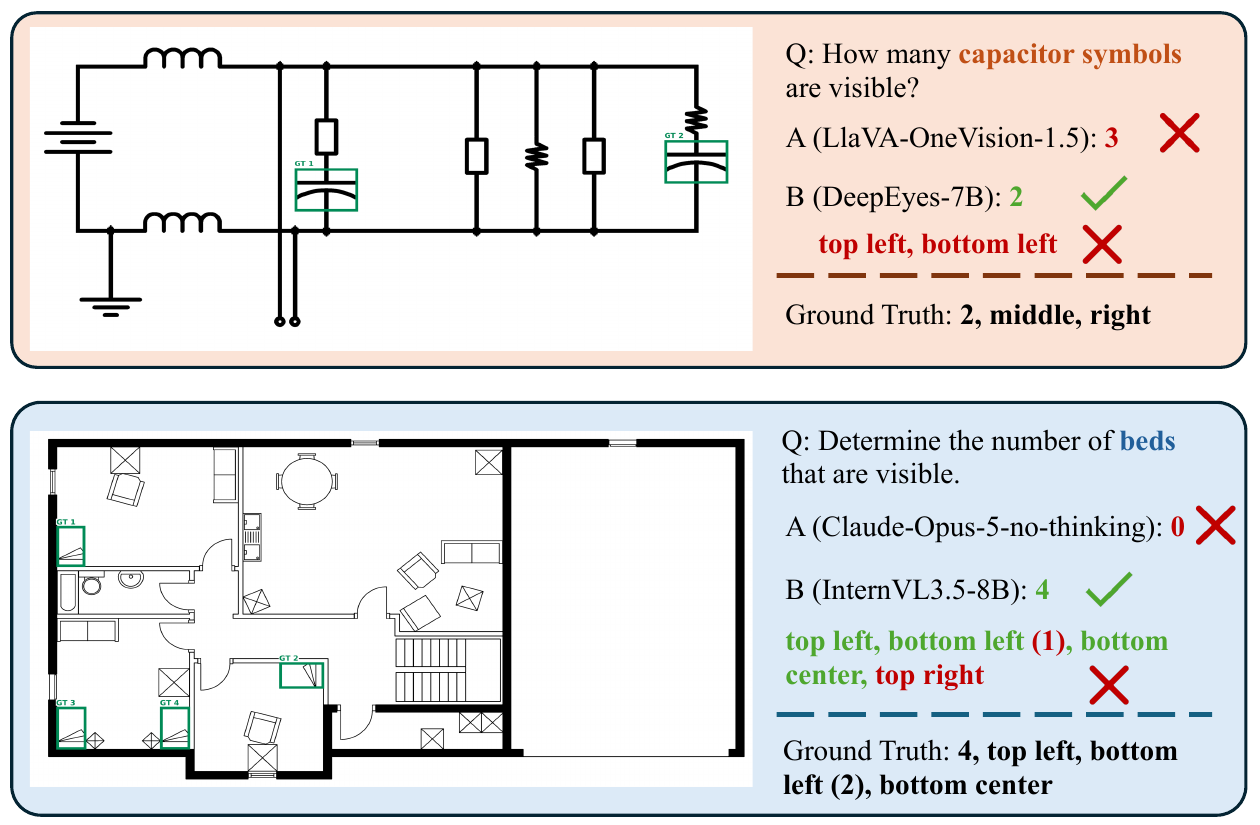}
    \caption{Some real cases of counting in technical drawings and motivation of our method. Even advanced MLLMs can fail in these tasks, and some MLLMs can provide the correct final number with wrong location of each instance.}
    \label{fig:teaser}
\end{figure}

Counting is a basic visual task for multi-modal large language models (MLLMs), which refers to determine how many instances are in the images. Previous works~\cite{paiss2023teaching,tamarapalli2025countqa,deitke2025molmo} have explored this area in natural images and propose insightful evaluation artifacts. However, when extending to another existing important domain, counting in technical drawings, we find it still largely underexplored, depicted in Figure~\ref{fig:teaser}. A door in a floor plan may appear as a short arc, circuit symbols can be small and visually similar, and carbon atoms in molecular structures are often implied by line endpoints. An MLLM must understand these drawing conventions, locate each target instance, and keep an accurate total across a dense image. These skills matter in engineering and scientific workflows, where counting is often a key step in later inspection and analysis.

Recent work has started to evaluate such topics. AECV-Bench studies counting and drawing QA on architectural plans~\cite{kondratenko2026aecv}, BlueprintSymVL evaluates symbol recognition in engineering blueprints~\cite{shteriyanov2025blueprintsymvl}. These efforts establish useful domain-specific testbeds. Based on these efforts, we argue that a common counting benchmark across different technical drawing types is needed, with instance-level references that support both evaluation and training.

To this end, we introduce \textbf{CountDrawBench}, a benchmark for counting objects in technical drawings. It contains 1,574 question-answer pairs from 707 images across six subsets collected from five source collections. The benchmark covers floor plans, circuit diagrams, mechanical drawings, and molecular structures. Instance bounding boxes are available for 722 questions. We construct the references from existing instance annotations, molecular graphs parsed with RDKit, and counts verified by four independent annotators for mechanical drawings. This design supports counting evaluation across different technical images. 

\noindent \textbf{A key observation.} As illustrated in Figure~\ref{fig:teaser}, in count-only evaluations, a correct answer may conceal incorrect visual grounding. As shown in Fig. 1, an MLLM can miss a target instance while counting an unrelated symbol, yet still return the correct total. This ambiguity is particularly important in technical drawings, where visually similar symbols often appear close together. It motivates evaluating not only how many objects are found, but also which objects are counted. 

Regarding the observation, we propose \textbf{Code-Verified RL}, a reinforcement learning framework that uses executable references to provide these training signals. Given a drawing and a counting question, the policy predicts the total count and, when reference boxes are available, a box for each predicted instance. A code interpreter checks the predicted count and evaluates one-to-one instance correspondence through bounding-box matching. We further introduce a local zoom recounting signal, which compares the full-image prediction with the sum of counts from four image tiles. These three signals are combined into a single GRPO objective~\cite{shao2024deepseekmath}, providing feedback on count accuracy, instance localization, and local visual consistency.

We train our model on the training set, CountDraw-7K, extended from our pipeline and conduct extensive experiments on CountDrawBench. Starting from Qwen3.5-9B, Code-Verified RL improves overall exact-count accuracy from 49.3\% to 54.3\%, outperforming the supervised fine-tuning baseline. We also evaluate the trained model on FSC-147 and CARPK to examine its transfer to natural-image counting.

Our contributions are summarized as follows:

\noindent \textbf{$\bullet$} \textbf{A benchmark for technical drawing counting}. We introduce CountDrawBench, covering six subsets across floor plans, circuit diagrams, mechanical drawings, and molecular structures, with 1,574 QA pairs and instance-level annotations for 722 questions.

\noindent \textbf{$\bullet$} \textbf{Executable references for counting and localization}. We establish a reference construction pipeline using existing instance annotations, molecular graph parsing, and independent human annotations. These references support both count evaluation and instance-level verification.

\noindent \textbf{$\bullet$} \textbf{Code-Verified RL for MLLM counting}. We develop a unified training objective combining executable count verification, one-to-one instance matching, and local zoom recounting. Experiments show consistent improvements over the Qwen3.5-9B baseline on technical drawing counting.

\section{Related Work}
\label{sec:related_work}

\textbf{Object Counting and Technical Diagram Understanding.}
It has been explored to estimate object totals on natural images via density regression or exemplar matching. Famous benchmarks like CountBench~\cite{paiss2023teaching}, PixMo-Count~\cite{deitke2025molmo}, and CountQA~\cite{tamarapalli2025countqa} provide natural-image counting benchmarks and point-grounded data.
Counting on technical images, instead, depends on drafting conventions, symbol geometry, and structural semantics.
Some representative tasks are symbol spotting in floor plans and schematics, construction and mechanical drawing QA, chemical structure recognition, and document layout analysis.
Regarding related works on this topic, AECV-Bench~\cite{kondratenko2026aecv} and BlueprintSymVL~\cite{shteriyanov2025blueprintsymvl} test symbol counting on engineering blueprints. DRAGON~\cite{iyengar2026dragon} tests whether models localize evidence boxes on diagrams given the gold answer.
Different from prior work, our benchmark unifies these technical families: floor plans, electrical schematics, mechanical drawings and chemical structures with RDKit-derived structural counts, into one benchmark with instance annotations or structure-derived count references. Table~\ref{tab:comparison} directly shows the differences.
\begin{table}[t]
\centering
\setlength{\tabcolsep}{4pt}
\renewcommand{\arraystretch}{1.2}
\resizebox{\linewidth}{!}{
\begin{tabular}{lcccccc}
\toprule2
\textbf{Benchmark} &
\shortstack{\textbf{Natural}\\\textbf{Images}} &
\shortstack{\textbf{Architecture}\\\textbf{Plans}} &
\shortstack{\textbf{Engineering}\\\textbf{Blueprints}} &
\shortstack{\textbf{Electrical}\\\textbf{Schematics}} &
\shortstack{\textbf{Mechanical}\\\textbf{Drawings}} &
\shortstack{\textbf{Molecular}\\\textbf{Structures}} \\
\midrule
\rowcolor{blue!5} CountBench~\cite{paiss2023teaching}      & \textbf{\checkmark} & & & & & \\
PixMo-Count~\cite{deitke2025molmo}          & \textbf{\checkmark} & & & & & \\
\rowcolor{blue!5} CountQA~\cite{tamarapalli2025countqa}            & \textbf{\checkmark} & & & & & \\
 FSC-147~\cite{ranjan2021learning} & \textbf{\checkmark} & & & & & \\
\rowcolor{blue!5} AECV-Bench~\cite{kondratenko2026aecv}       & & \textbf{\checkmark} & & & & \\
 BlueprintSymVL~\cite{shteriyanov2025blueprintsymvl} & & & \textbf{\checkmark} & & & \\
\midrule
\rowcolor{red!5} \textbf{CountDrawBench (Ours)} &
& \textbf{\checkmark} & & \textbf{\checkmark} & \textbf{\checkmark} & \textbf{\checkmark} \\
\bottomrule
\end{tabular}
}
\caption{
Domain coverage of representative visual counting benchmarks.
CountDrawBench extends counting evaluation to multiple technical
drawing domains.
}
\label{tab:comparison}
\end{table}

\noindent \textbf{Policy Optimization for Visual Perception.} Several previous works~\cite{liu2025visual,shen2025vlm,wang2026traceable} combine RL-based policy optimization methods such as GRPO~\cite{xu2026deepseek} and visual localization with box-IoU rewards. Methods like ZoomEye~\cite{shen2025zoomeye} and DeepEyes~\cite{zheng2026deepeyes} further apply visual tool calling to help perception tasks.
More recent works apply a new technique: On-policy self-distillation (OPSD), which instead aligns teacher and student token distributions along student rollouts under asymmetric visual inputs. Vision-OPD~\cite{yuan2026vision} supervises a full-image policy with crop-conditioned predictions, and LT-OPD~\cite{li2026fewer} distills full-vision teacher distributions under extreme visual-token reduction.

\noindent \textbf{Tool-Augmented Vision and Code-Verified Rewards.}
Visual search and programmatic frameworks equip MLLMs with guided cropping and Python execution, while RL-trained agents invoke zoom-in actions or image-manipulation code during reasoning.
CodeV~\cite{hou2026codev} shows that a correct final answer can mask unfaithful tool calls on irrelevant regions, and combines outcome rewards with VLM-judged process rewards over intermediate tool outputs.
Inspired by prior work, we execute a code interpreter to check the count, measure bounding-box IoU, and compare the full-image answer to a zoomed-crop answer within one GRPO objective.

\begin{figure*}[t]
    \centering
    \includegraphics[width=\linewidth]{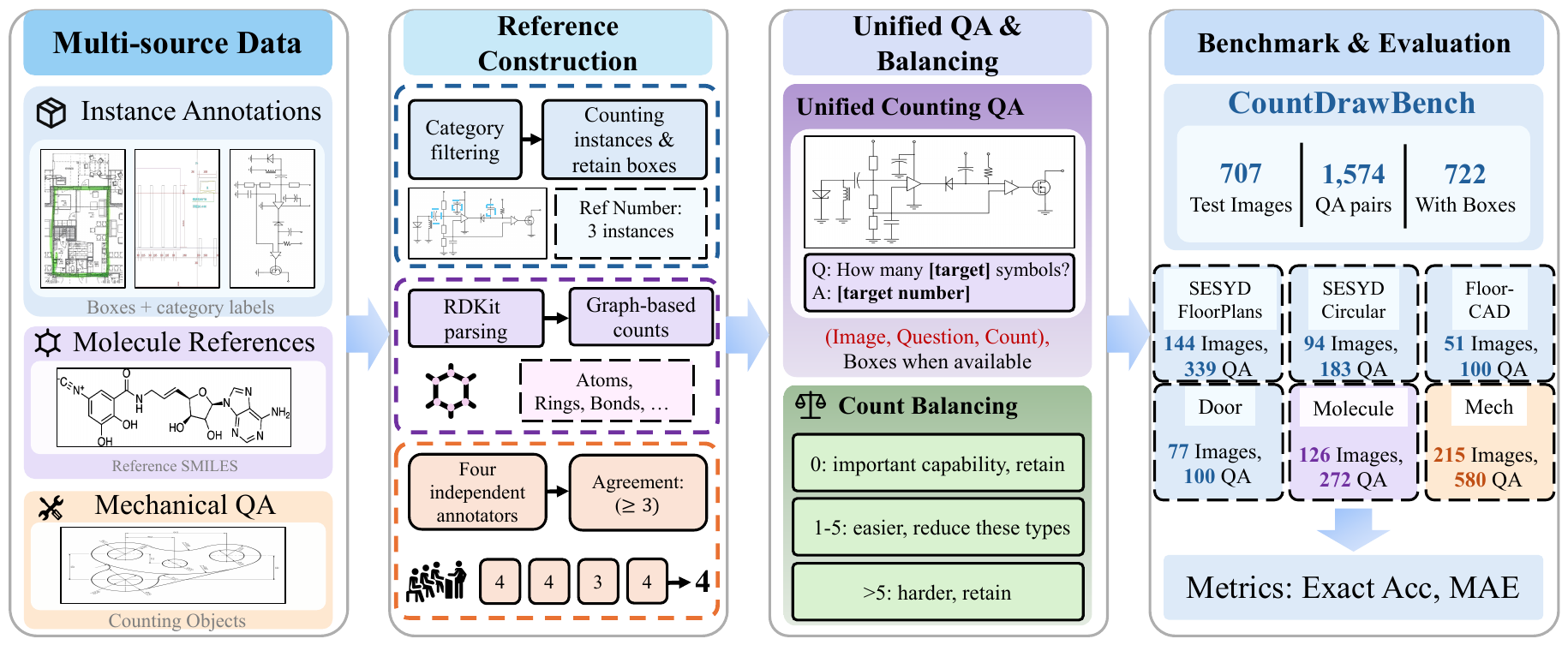}
    \caption{Overview of CountDrawBench and our proposed pipeline. The whole process contains four steps: source data selection, reference construction, unified QA construction\& question balancing and evaluation. }
    \vspace{-3mm}
    \label{fig:pipeline}
\end{figure*}

\section{CountDrawBench}
\label{sec:count_draw}

CountDrawBench evaluates how well a model can count a target named in a question about a technical drawing.
We select these representative dimensions: floor and door plans, circuit diagrams, mechanical drawings, and molecular diagrams, and then build the benchmark. Details can be found in Figure~\ref{fig:pipeline}.
These images use different drawing conventions to represent objects and structures.
A door may appear as a short arc, while a carbon atom may be implied by a line endpoint.
The task therefore calls for both reading the drawing and keeping track of repeated instances.

Each example contains an image, an English counting question, and a nonnegative integer answer. The question specifies the target, such as a capacitor symbol, a bedroom door, an orthographic view, or a nitrogen atom. Where instance annotations are available, we also retain bounding boxes for the queried category. These boxes support training and allow us to assess the visual evidence behind a predicted count. 

\subsection{Data sources and construction}
We assemble 1,574 question-answer pairs from five source collections, with SESYD~\cite{delalandre2010generation} contributing both floorplans and circuit diagrams.
Figure~\ref{fig:pipeline} summarizes the six source subsets.
The labels come from three routes: existing instance annotations, reference molecular structures, and mechanical counting questions.

\noindent \textbf{Counts from instance annotations.}
SESYD~\cite{delalandre2010generation} and FloorPlanCAD~\cite{fan2021floorplancad} (Floor-CAD) provide labeled symbol instances.
For each selected category, we count its instances and retain their boxes.
In SESYD, shape variants of the same symbol are grouped under a common name, so the question refers to a category such as doors or capacitors.
CubiCasa5K~\cite{kalervo2019cubicasa5k} (Door) supplies room-related door categories, such as bedroom doors and kitchen doors.
We preserve these category definitions and count the corresponding annotations. 

\noindent \textbf{Counts from molecular structures.}
We use the molecular images and reference structures in the ChemOCR data associated with ChemVLM~\cite{li2025chemvlm}.
Each reference structure gives a molecular graph from which we derive counts for eight atom types, rings, aromatic rings, and double bonds.
The question asks about one of these targets in the paired image.
Zero-count examples are retained so that the collection also tests whether a requested atom or structure is absent.
The reference graph accounts for atoms implied by the drawing convention, including unlabeled carbon atoms. We use ``Molecule'' to represent this subset. 

\noindent \textbf{Counts in mechanical drawings.}
We first collect the counting questions in MechVQA~\cite{kou2026mechvqa}. To extend this subset, we conduct a human labeling process of objects in the images. Four graduate students are recruited to do this job.  
A target is retained when it appears in at least three people's labeling results.
We keep a count directly when at least three passes give the same number.
This yields the full generated questions: a set including original data with ground truth counts and extended data with human-labeled object counts.
We keep the label origin and agreement level for each question. We use ``Mech'' to represent this subset.

\subsection{Dataset coverage}
We first reduce the share of questions with counts from one to five.
Within each source collection, we cap their number at three times the number of questions (much more than 3x originally) with counts of six or more.
Zero-count questions are kept separately. After this process, CountDrawBench is built successfully. This benchmark set contains 108 zero-count questions, 944 questions with answers from one to five, and 522 with answers above five.
Test counts range from zero to 76.
Bounding boxes are available for 722 test questions.
Together, the integer labels and instance boxes support evaluating both count accuracy and localization.

\subsection{Evaluation}
\label{sec:eval}
We use exact count accuracy as the primary metric and report mean absolute error (MAE) to show the size of counting errors. The higher count accuracy and the lower MAE represents better performances. 
We also report results by source collection and count range.
Responses without a valid integer answer count as incorrect. 
We report their rate and use zero for their count-error calculation. Additionally, we report mIoU on the 722 questions whose bounding boxes are available, to show localization capabilities of models. The mIoU scores are per-question averaged, and missing or invalid predicted bounding boxes are assigned an IoU of zero.

\section{Our Method: Code-Verified RL}
\label{sec:code_verified_rl}

\begin{figure}[t]
    \centering
    \includegraphics[width=\linewidth]{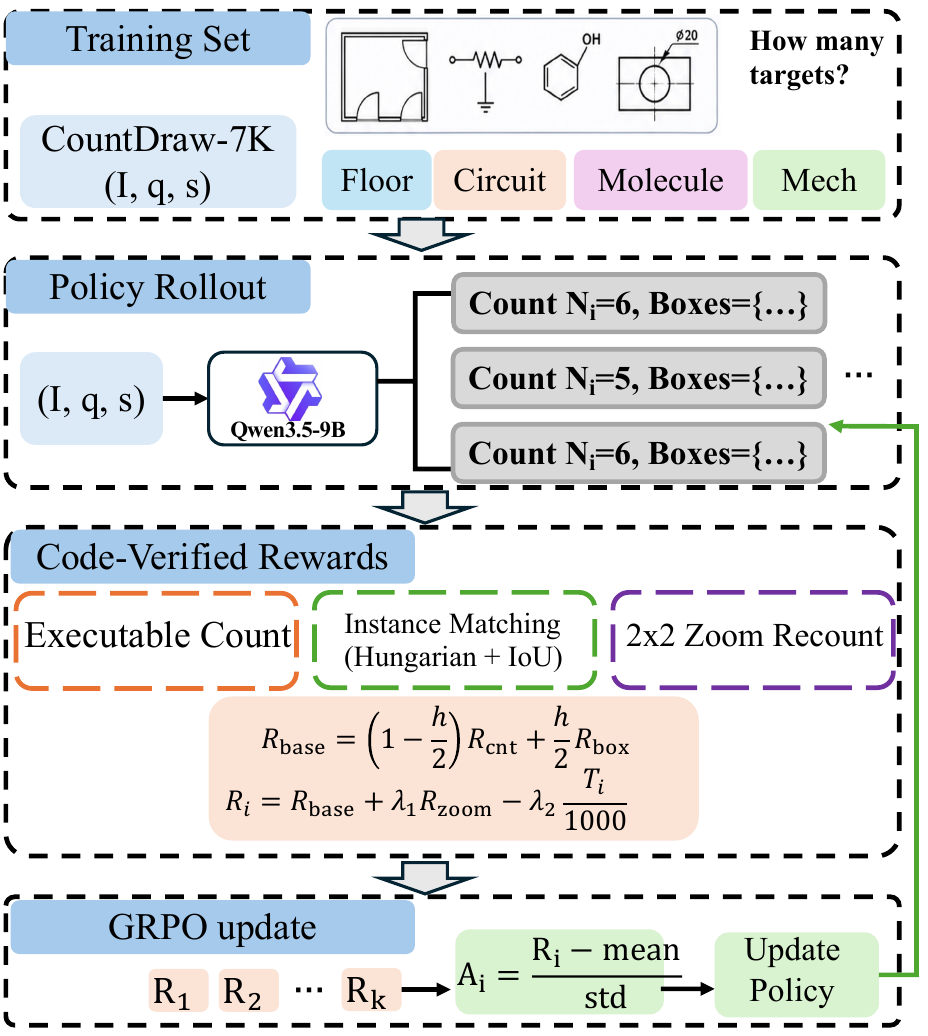}
    \caption{Our Training Pipeline. The code-verified RL is trained on newly constructed data, with rewards from executable counting, instance matching and zoom recounting.}
    \vspace{-3mm}
    \label{fig:rl}
\end{figure}
Counting in a technical drawing requires identifying the target instances and keeping track of their total. These two operations can fail in ways that conceal each other: omitting one door and including one unrelated symbol leaves the final count unchanged. Drawing conventions add a second challenge. A short arc can denote a door, while a line endpoint can denote an unlabeled carbon atom. To solve this problem, we extend the pipeline in Section~\ref{sec:count_draw} and construct a training set. The training labels provide complementary ways to supervise these decisions: integer answers specify cardinality, instance boxes specify spatial correspondence, and molecular structures specify counts.

We use these references to train an MLLM policy through \textbf{Code-Verified RL}. As shown in Figure~\ref{fig:rl}, the policy generates an answer, and executable checks assign rewards to its count and, where annotated, its instance boxes. A local recount supplies an additional comparison at a finer visual scale. Group-relative optimization then increases the probability of responses with stronger combined evidence.
\subsection{Training Data Construction}
\label{sec:training_data}
We extend the data creation pipeline described in Section~\ref{sec:count_draw} and create a larger training set, \textbf{CountDraw-7K}, for our policy optimization process. We have carefully checked the training set and the benchmark to avoid any data leakage or image overlap. Each piece of data in CountDraw-7K also contains normal VQA pairs. The distributions are: Floor-CAD 1,823; SESYD 2,155; Mech 1,449; Molecule 1,154 and Door 419. We apply all our training based on it. 

\subsection{Counting with executable references}
Let a training example be $(I,q,s)$, where $I$ is an image, $q$ is the counting target, and $s$ contains the reference annotations described in Section~\ref{sec:count_draw}. The reference count $N^*$ is
\begin{equation}
\label{eq:cvrl_reference}
N^*=v(s,q),\qquad D=\max(N^*,1),
\end{equation}
where $D$ is a factor for calculating reward. For molecular questions, $v$ parses the reference SMILES with RDKit and executes the query for the requested atom type, rings, aromatic rings, or double bonds. Carbon counts therefore include atoms implied by skeletal notation. For the other sources, $v$ reads the integer label obtained from instance annotations or mechanical counting questions. This connects the benchmark's label construction to the training feedback.

Given $(I,q)$, the policy produces a response $y_i$ whose final answer is a nonnegative integer $N_i$. For questions with compatible reference boxes, we also request one box per counted instance. The boxes use coordinates normalized to $[0,1000]$ on the original image. This representation associates the requested target with explicit locations, a form of supervision used in language-conditioned counting~\cite{cheng2025pointarena} and grounded multimodal models~\cite{lai2024lisa}. The verifier evaluates the parsed answer and geometry against the references.

\subsection{Verifying cardinality and instance matching}
\label{sec:verifying}

We assign the count reward
\begin{align}
\label{eq:cvrl_count}
& R_{\mathrm{cnt}}(N_i,N^*)= \nonumber \\
& \begin{cases}
1, & N_i=N^*,\\
\frac{1}{2}\max\!\left(0,1-\frac{|N_i-N^*|}{D}\right),
& \text{valid }N_i\ne N^*,\\
0, & \text{otherwise}.
\end{cases}
\end{align}
Exact answers receive a distinct maximum, while nearby counts receive partial credit according to relative error. Missing or invalid answers receive zero. The denominator accommodates the benchmark's range of counts and its zero-count questions.

For annotated examples, let $B_i=\{b_j\}_{j=1}^{m_i}$ be the predicted boxes and $B^*=\{b_k^*\}_{k=1}^{m^*}$ the reference boxes. We compute a maximum-overlap one-to-one matching $M^*$ with the Hungarian algorithm and use
\begin{equation}
\label{eq:cvrl_boxes}
\begin{aligned}
M^*&=\underset{M\in\mathcal{M}(B_i,B^*)}{\arg\max}
\sum_{(j,k)\in M}\operatorname{IoU}(b_j,b_k^*),\\
R_{\mathrm{box}}&=
\frac{\sum_{(j,k)\in M^*}\operatorname{IoU}(b_j,b_k^*)}
{\max(m_i,m^*)}.
\end{aligned}
\end{equation}
Here $\mathcal{M}$ denotes the set of one-to-one matchings. Unmatched instances contribute zero, so both omissions and extra predictions lower the reward. Each parsed box must satisfy $0\le x_1<x_2\le1000$ and $0\le y_1<y_2\le1000$, and $m_i$ must equal $N_i$. We set $R_{\mathrm{box}}=0$ for an empty prediction or a failed count, cardinality, or geometry check. Thus, two responses with the same integer can receive different rewards according to which instances they identify.

Let $h=1$ when compatible, nonempty reference boxes are available and $h=0$ otherwise. The reference-based reward is
\begin{equation}
\label{eq:cvrl_base}
R_{\mathrm{base}}=
\left(1-\frac{h}{2}\right)R_{\mathrm{cnt}}
+\frac{h}{2}R_{\mathrm{box}}.
\end{equation}
On boxed examples, cardinality and correspondence receive equal weight. On the remaining examples, the full weight goes to cardinality. This keeps the attainable maximum at one across the benchmark's annotation regimes, while using the spatial supervision available for architectural plans and circuit diagrams. Equation~\eqref{eq:cvrl_boxes} also follows the matched-IoU definition used in Section~\ref{sec:count_draw}, linking training feedback to localization evaluation.

\subsection{Recounting at a finer visual scale}

Small repeated symbols motivate examining local regions alongside the whole drawing. For each training question, we divide the image into a fixed, disjoint $2\times2$ grid and resize each tile for the vision encoder. The current policy greedily counts the requested target in each tile using a short prompt and a budget of 128 tokens. The prompt asks for recognizable target instances within that tile. From the four parsed tile counts, we form $N_{\mathrm{zoom}}=\sum_{r=1}^{4}n_r$ and assign
\begin{equation}
R_{\mathrm{zoom}}
= -\frac{1}{2}\frac{|N_{\mathrm{zoom}}-N_i|}{D}
+\frac{1}{2}\mathbf{1}[N_{\mathrm{zoom}}=N^*].
\label{eq:zoom_reward}
\end{equation}
If the main count or any tile count is missing, this
component is zero. Tile responses are generated without
gradients and reused for repeated candidates within the
current reward batch; they are recomputed as training
progresses.

The local recount supplies a shared estimate against
which candidate whole-image answers can be compared.
When all candidates have valid counts and share
$N_{\mathrm{zoom}}$, its bonus from reference is constant
and cancels during group centering. The
candidate-dependent signal comes from the disagreement
term. The separate count reward anchors this comparison
to $N^*$. This gives local visual agreement and reference
correctness distinct roles in selecting responses.

For a main response of $T_i$ generated tokens, the complete reward is
\begin{equation}
\label{eq:cvrl_reward}
R_i=R_{\mathrm{base}}+\lambda_{1} R_{\mathrm{zoom}}
-\lambda_{2}\frac{T_i}{1000},
\end{equation}
where $\lambda_{1}, \lambda_{2}$ are hyperparameters, setting to 0.3 and 0.01 empirically. The token cost (per 1000 tokens) encourages concise responses while retaining room for instance enumeration. It applies to the main response, including its generated end token.

\begin{table*}[t]
  \centering
  \caption{Main Results. We use \textbf{\textcolor{red}{red}} to represent the best performance of open-source models, and \textbf{\textcolor{blue}{blue}} to represent the second best one.}
  \label{tab:countdraw-main}
  \begingroup
  \small
  \setlength{\tabcolsep}{3pt}
  \setlength{\fboxsep}{2pt}
  \renewcommand{\arraystretch}{1.24}
  \newcommand{\CountDrawHead}[1]{%
    \raisebox{-6.3pt}[0pt][0pt]{\textnormal{#1}}}
  \newcommand{\CountDrawMAE}[1]{\makebox[2.5em][r]{#1}}
  \newcommand{\CountDrawGroup}[2]{%
    \multicolumn{10}{@{}l@{}}{%
      \makebox[0pt][l]{\colorbox{#1}{%
        \makebox[\dimexpr\textwidth-2\fboxsep\relax][c]{%
          \strut\textnormal{#2}}}}}}
  \begin{tabular*}{\textwidth}{@{\extracolsep{\fill}}lccccccccc@{}}
    \toprule
    \CountDrawHead{Model} & \CountDrawHead{Params}
      & \multicolumn{6}{c}{\textnormal{Count accuracy (\%) $\uparrow$}}
      & \CountDrawHead{MAE $\downarrow$} & \CountDrawHead{mIoU $\uparrow$} \\
    \cmidrule(lr){3-8}
    & & \textnormal{SESYD} & \textnormal{Floor-CAD} & \textnormal{Door}
      & \textnormal{Molecule} & \textnormal{Mech} & \textnormal{Overall} &  \\
    \midrule
    \CountDrawGroup{gray!8}{\textup{Proprietary MLLMs}} \\
    \addlinespace[3pt]
    GPT-5.6-sol (effort=medium)~\cite{singh2025openai} & --- & 57.0 & 26.0 & 41.0 & 86.1 & 43.5 & 54.1 & \CountDrawMAE{1.39} & \CountDrawMAE{42.51} \\
Claude-Opus-5 (no-think)~\cite{claude_opus5} & --- & 47.9 & 25.0 & 34.0 & 91.2 & 44.7 & 51.9 & \CountDrawMAE{1.49} & \CountDrawMAE{29.01} \\
    Gemini-3.8-Flash~\cite{gemini3_8}      &  ---    & 46.0 & 17.0 & 39.0 & 89.5 & 67.8 & \textnormal{59.3} & \CountDrawMAE{2.93} & \CountDrawMAE{30.22} \\
    \addlinespace[4pt]
    \CountDrawGroup{blue!8}{\textup{Open-weight MLLMs}} \\
    \addlinespace[3pt]
    Qwen3-VL (Instruct)~\cite{bai2025qwen3} & 8B   & 16.9 & 24.0 & 28.0 & 43.4 & 20.7 & \textnormal{24.0} & \CountDrawMAE{9.95} & \CountDrawMAE{0.08} \\
    Qwen3.5-9B~\cite{bai2025qwen3}  & 9B & 32.8 & 12.0 & 20.0 & \textcolor{blue}{\textbf{70.3}} & \textbf{\textcolor{blue}{65.7}} & \textnormal{49.3} & \CountDrawMAE{3.37} & \CountDrawMAE{8.92} \\
    InternVL3.5-HF~\cite{wang2025internvl3}   & 8B   & 13.6 & \textcolor{blue}{\textbf{22.0}} & 24.0 & 58.8 & 16.7 & \textnormal{23.8} & \CountDrawMAE{14.85} & \CountDrawMAE{10.09} \\
    GLM-4.6V-Flash~\cite{hong2025glm}      & 9B   & 18.8 & \textbf{\textcolor{red}{30.0}} & \textbf{\textcolor{red}{36.0}} & 37.9 & 20.7 & \textnormal{24.6} & \CountDrawMAE{14.73} & \CountDrawMAE{0.08} \\
    DeepSeek-V4.1-Flash~\cite{xu2026deepseek}  & 552B & \textbf{\textcolor{blue}{33.0}} & 17.0 & \textbf{\textcolor{blue}{28.0}} & \textcolor{red}{\textbf{81.8}} & 63.1 & \textcolor{blue}{\textbf{51.2}} & \textcolor{blue}{\textbf{\CountDrawMAE{2.81}}} & \CountDrawMAE{8.33} \\
    
    \addlinespace[4pt]
    \CountDrawGroup{green!8}{Tool-augmented MLLMs} \\
    \rule{0pt}{4.2mm} Qwen3-VL + ZoomEye~\cite{shen2025zoomeye} (EMNLP'25)  & 8B   & 22.7 & 23.0 & 24.0 & 55.5 & 31.0 & 31.5 & \CountDrawMAE{10.52}  & \CountDrawMAE{6.04}\\
 \rule{0pt}{4.2mm} DeepEyes~\cite{zheng2026deepeyes} (ICLR'26) & 7B & 23.2 & 17.0 & 22.0 & 58.1 & 24.7 & 29.3 &  \CountDrawMAE{3.30} & \CountDrawMAE{1.09}  \\
    \rule{0pt}{4.2mm} Vision-OPD~\cite{yuan2026vision} (NeurIPS'26) & 9B & 27.5 & 14.0 & 23.0 & 69.3 & 64.9 & 47.4 & \CountDrawMAE{3.44} & \CountDrawMAE{8.33} \\
    \addlinespace[2pt]
    \CountDrawGroup{orange!8}{\textnormal{Ours}} \\
    \rule{0pt}{4.2mm} Qwen3.5 + SFT & 9B & 37.6 & 14.0 & 21.0 & 63.5 & 56.4 & 46.4 & \CountDrawMAE{3.02} & \textbf{\textcolor{blue}{\CountDrawMAE{16.48}}} \\
    \rule{0pt}{4.2mm} \textbf{CountDraw} & 9B & \textbf{\textcolor{red}{47.2}} & 14.0 & 23.0 & 67.6 & \textbf{\textcolor{red}{66.8}} & \textcolor{red}{\textbf{54.3}} & \textcolor{red}{\textbf{\CountDrawMAE{2.80}}} & \textcolor{red}{\textbf{\CountDrawMAE{35.83}}} \\
    \bottomrule
  \end{tabular*}
  \endgroup
  \vspace{-3mm}
\end{table*}

\subsection{Optimization Objective}

We initialize the policy from Qwen3.5-9B and update all parameters on the 7,000 training questions from Section~\ref{sec:training_data}. For each question, we sample $K=8$ main responses at temperature 1 and compute group-relative advantages \cite{shao2024deepseekmath}:
\begin{equation}
\label{eq:cvrl_advantage}
\begin{aligned}
\bar R&=\frac{1}{K}\sum_{j=1}^{K}R_j,\\
\sigma_R&=\sqrt{\frac{\sum_{j=1}^{K}(R_j-\bar R)^2}{K-1}},\\
A_i&=\frac{R_i-\bar R}{\sigma_R+10^{-4}}.
\end{aligned}
\end{equation}
Each generated token inherits its response's advantage. We adopt the fixed-length loss aggregation introduced by Dr.GRPO~\cite{liu2025understanding} and retain within-group standard-deviation scaling. For token context $z_{it}=(I,q,y_{i,<t})$, write $\rho_{it}=\pi_\theta(y_{it}\mid z_{it})/\pi_{\mathrm{old}}(y_{it}\mid z_{it})$ and $\tilde\rho_{it}=\operatorname{clip}(\rho_{it},1-\epsilon,1+\epsilon)$. We maximize
\begin{equation}
\label{eq:cvrl_objective}
\begin{aligned}
\mathcal J(\theta)
&=\mathbb E\!\left[
\frac{1}{K L_{\max}}
\sum_{i=1}^{K}\sum_{t=1}^{T_i}
\left(c_{it}S_{it}-\beta d_{it}\right)
\right],\\
S_{it}&=\min\!\left(\rho_{it}A_i,\tilde\rho_{it}A_i\right).
\end{aligned}
\end{equation}
The resulting training signal follows the benchmark's two evaluation axes: correct totals and matched instance evidence. Source-wise evaluation tests how these decisions transfer to held-out drawings with different conventions, while count-range evaluation measures how well the policy tracks repeated instances as their number grows.

\section{Experiments}
\label{sec:experiments}

\subsection{Experimental Settings}
\label{sec:exp_settings}

\paragraph{Models and baselines.}
We apply our method on an advanced MLLM, Qwen3.5-9B~\cite{bai2025qwen3} and compare the trained model (\textbf{CountDraw-9B}) against both general MLLMs and specialized perception baselines.
For general MLLMs, we choose leading proprietary models including GPT-5.6-sol~\cite{singh2025openai}, Claude-Opus-5, and Gemini-3.8-Flash, as well as open-weight models including Qwen3-VL-8B-Instruct, Qwen3.5-9B, InternVL3.5-8B-HF~\cite{wang2025internvl3}, GLM-4.6V-Flash~\cite{hong2025glm}, and DeepSeek-V4.1-Flash~\cite{xu2026deepseek}. All proprietary models and DeepSeek-V4.1-Flash are called by API. 
We also compare with zoom-in and fine-grained perception methods, including ZoomEye~\cite{shen2025zoomeye} applied to Qwen3-VL-8B, the RL-trained agentic model DeepEyes-7B~\cite{zheng2026deepeyes}, and the self-distilled model Vision-OPD-9B~\cite{yuan2026vision}.
Finally, to isolate the effect of our code-verified RL training on CountDraw-7k, we include supervised fine-tuning (SFT) on the same Qwen3.5-9B backbone as our training baseline.

\noindent \textbf{Basic settings.}
During evaluation, models receive the input image and question at temperature $0$, and the other hyperparameters follow a default setting.
Thinking mode is disabled for all proprietary MLLMs, while GPT-5.6-sol is evaluated with medium reasoning effort.

\noindent \textbf{Datasets and benchmarks.}
All models are evaluated on CountDrawBench, which contains 1,574 questions.
Additionally, we add two external counting benchmarks: FSC-147 and CARPK, to show whether our trained CountDraw-9B can be generalized to counting tasks in other domains.

\noindent \textbf{Evaluation metrics.}
The evaluation metrics are listed in Section~\ref{sec:eval}. All evaluated models follow the same metrics.

\begin{table*}[t]
  \centering
  \caption{Performance across count ranges. We have four groups: 0 instances, 1-5 instances, 6-10 and 11+ instances. We use \textbf{\textcolor{red}{red}} to represent the best performance of open-source models, and \textbf{\textcolor{blue}{blue}} to represent the second best one.}
  \label{tab:object-count-countdrawbench}
  \begingroup
  \small
  \setlength{\tabcolsep}{3pt}
  \setlength{\fboxsep}{1.9pt}
  \renewcommand{\arraystretch}{1.16}

  \newcommand{\CountDrawCountGroup}[1]{%
    \multicolumn{9}{@{}l@{}}{%
      \makebox[0pt][l]{\colorbox{gray!8}{%
        \makebox[\dimexpr\textwidth-2\fboxsep\relax][c]{\strut #1}}}}}

  \newcommand{\CountDrawCountShade}{%
    \makebox[0pt][l]{\raisebox{-1.16\dp\strutbox}[0pt][0pt]{%
      \textcolor{orange!10}{\rule{\textwidth}{%
        \dimexpr1.16\ht\strutbox+1.16\dp\strutbox\relax}}}}}

  \begin{tabular*}{\textwidth}{@{\extracolsep{\fill}}l*{8}{r}@{}}
    \toprule
    CountDrawBench
      & \multicolumn{2}{c}{0 instances}
      & \multicolumn{2}{c}{1--5 instances}
      & \multicolumn{2}{c}{6--10 instances}
      & \multicolumn{2}{c}{11+ instances} \\
    \cmidrule(lr){2-3}
    \cmidrule(lr){4-5}
    \cmidrule(lr){6-7}
    \cmidrule(lr){8-9}

    Model / setting
      & Acc.\ (\%) $\uparrow$ & MAE $\downarrow$
      & Acc.\ (\%) $\uparrow$ & MAE $\downarrow$
      & Acc.\ (\%) $\uparrow$ & MAE $\downarrow$
      & Acc.\ (\%) $\uparrow$ & MAE $\downarrow$ \\
    \midrule

    \CountDrawCountGroup{Baseline models} \\
    \addlinespace[1pt]

    Gemini-3.8-Flash~\cite{gemini3_8}
      & 91.7 & 2.23
      & 57.8 & 1.86
      & 51.8 & 4.62
      & 61.8 & 5.70 \\

    Qwen3.5-9B~\cite{bai2025qwen3}
      & 88.0 & 3.31
      & 48.5 & 2.75
      & 37.5 & 3.80
      & \textcolor{blue}{\textbf{52.2}} & 5.81 \\

    DeepSeek-V4.1-Flash~\cite{xu2026deepseek}
      & \textcolor{red}{\textbf{90.7}} & \textcolor{blue}{\textbf{2.28}}
      & \textcolor{red}{\textbf{51.9}} & \textcolor{red}{\textbf{2.05}}
      & 36.0 & 3.37
      & \textcolor{blue}{\textbf{52.2}} & 5.93 \\

    Vision-OPD-9B~\cite{yuan2026vision}
      & 87.0 & 3.56
      & 45.7 & 3.18
      & \textcolor{blue}{\textbf{37.8}} & \textcolor{blue}{\textbf{2.83}}
      & 50.5 & \textcolor{blue}{\textbf{5.74}} \\

    \addlinespace[2.5pt]
    \CountDrawCountGroup{Ours} \\
    \addlinespace[1pt]

    \CountDrawCountShade CountDraw-9B
      & \textcolor{blue}{\textbf{88.9}} & \textcolor{red}{\textbf{1.77}}
      & \textcolor{blue}{\textbf{51.8}} & \textcolor{blue}{\textbf{2.65}}
      & \textcolor{red}{\textbf{49.4}} & \textcolor{red}{\textbf{2.19}}
      & \textbf{\textcolor{red}{55.9}} & \textcolor{red}{\textbf{5.25}} \\

    \bottomrule
  \end{tabular*}
  \endgroup
  \vspace{-3mm}
\end{table*}

\subsection{Main Results}
\label{sec:main_res}
Table~\ref{tab:countdraw-main} presents the overall performance on CountDrawBench. We highlight these key observations:

First, counting in technical drawings remains \textbf{challenging} for current MLLMs, particularly open-source models. Even strong models such as Qwen3.5-9B and DeepSeek-V4.1-Flash achieve only 49.3\% and 51.2\% accuracy, respectively. Proprietary models also show considerable room for improvement, with the best accuracy reaching 59.3\%. Additionally, model performances are \textbf{inconsistent} across sub-tasks. Sub-tasks like Floor-CAD and Door are much harder, as all models perform not so well. Many MLLMs achieve only a 10\%-20\% accuracy.

More notably, our CountDraw-9B achieves 54.3\% accuracy, improving its Qwen3.5-9B backbone by 5 percentage points and outperforming all evaluated open-source baselines. It also reduces MAE from 3.37 to 2.80. The model even outperforms DeepSeek-V4.1-Flash, an MLLM with a huge amount of parameters, in cases with 6+ instances. 

Besides accuracy, we observe a larger \textbf{improvement in instance localization}. CountDraw increases mIoU from 8.92 to 35.83, alongside a 5.0-point gain in counting accuracy. This suggests that the benefit extends beyond predicting the correct total: the model also becomes better at identifying the corresponding instances. The result supports our motivation that count correctness and visual grounding should be considered together.

\subsection{Count-Level Results}
\label{sec:count_level}

\textbf{Performance across different count ranges.} Table 3 presents the results across four count ranges. We find that a clear challenge emerges when the target count increases to 6-10. Even Gemini-3.8-Flash achieves only 51.8\% accuracy on 6-10, while open-source baselines remain below 38\%. Our CountDraw-9B achieves 49.4\% accuracy in this range, improving upon its backbone by 11.9 percentage points and reducing MAE from 3.80 to 2.19. The gains also extend to larger counts (11+), where CountDraw-9B achieves the best accuracy (55.9\%) and lowest MAE (5.25) among open-source models. On zero-count questions, our model further achieves the lowest MAE of 1.77 across all evaluated models. These results show that CountDraw-9B improves counting across different count ranges, with particularly strong gains on challenging counting questions.

\begin{figure}[t]
    \centering
    \includegraphics[width=\columnwidth]{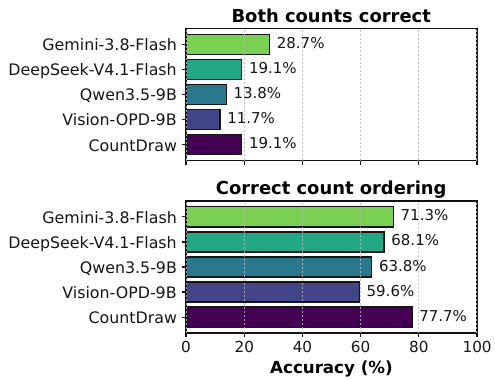}
    \caption{Exact-count and relative-order accuracy for pairs of questions about different targets in the same image.}
    \label{fig:same_image_counting}
    \vspace{-5mm}
\end{figure}

\noindent\textbf{Performance when changing targets.}
We pair questions about different targets with unequal reference counts in the same image, yielding 94 pairs across 86 images from CountDrawBench.
CountDraw-9B preserves the reference count order on 77.7\% of pairs, compared with 63.8\% for Qwen3.5-9B (Figure~\ref{fig:same_image_counting}).
CountDraw answers both questions correctly on 19.1\% of pairs, and Gemini-3.8-Flash leads on this metric at 28.7\%.

\begin{table}[t]
  \centering
  \caption{Ablation study of the IoU and zoom-consistency rewards on CountDrawBench. We \textbf{bold} the best scores.}
  \vspace{-2mm}
  \label{tab:countdraw-ablation}
  \begingroup
  \small
  \setlength{\tabcolsep}{5pt}
  \renewcommand{\arraystretch}{1.18}
  \newcommand{\CountDrawAblationShade}{%
    \makebox[0pt][l]{%
      \raisebox{-1.18\dp\strutbox}[0pt][0pt]{%
        \textcolor{orange!10}{\rule{\columnwidth}{%
          \dimexpr1.18\ht\strutbox+1.18\dp\strutbox\relax}}}}}
  \begin{tabular*}{\columnwidth}
    {@{\extracolsep{\fill}}lrr@{}}
    \toprule
    \textbf{Method}
    & \textbf{Acc. (\%) $\uparrow$}
    & \textbf{MAE $\downarrow$} \\
    \midrule
    \CountDrawAblationShade Full Code-Verified RL
      & \textbf{54.3} & \textbf{2.80} \\
    w/o $R_{\mathrm{box}}$
      & 52.4 & 3.03 \\
    w/o $R_{\mathrm{zoom}}$
      & 50.3 & \textbf{2.80} \\
    \bottomrule
  \end{tabular*}
  \endgroup
  \vspace{-2mm}
\end{table}

\begin{table*}[t]
  \centering
  \caption{Results on two external counting benchmarks: FSC-147 and CARPK.
  Following the original settings, we also report RMSE besides MAE.
  \textbf{\textcolor{red}{Red}} and
  \textbf{\textcolor{blue}{blue}} indicate the best and second-best
  performance among open-source models, respectively.}
  \vspace{-1mm}
  \label{tab:countdraw-external}

  \begingroup
  \small
  \setlength{\tabcolsep}{5pt}
  \renewcommand{\arraystretch}{1.16}

  \begin{tabular*}{\textwidth}{@{\extracolsep{\fill}}lccccccc@{}}
    \toprule
    \textbf{Model} & \textbf{Params}
      & \multicolumn{3}{c}{\textbf{FSC-147}}
      & \multicolumn{3}{c}{\textbf{CARPK}} \\
    \cmidrule(lr){3-5}\cmidrule(lr){6-8}
    & & Acc. (\%) $\uparrow$ & MAE $\downarrow$ & RMSE $\downarrow$
      & Acc. (\%) $\uparrow$ & MAE $\downarrow$ & RMSE $\downarrow$ \\
    \midrule

    \multicolumn{8}{c}{\cellcolor{gray!8}\textit{Proprietary MLLMs}} \\
    Gemini-3.8-Flash~\cite{gemini3_8} & ---
      & 35.4 & 16.15 & 134.47
      & 4.5 & 21.29 & 25.03 \\

    \addlinespace[3pt]
    \multicolumn{8}{c}{\cellcolor{blue!8}\textit{Open-weight MLLMs}} \\

    Qwen3-VL (Instruct)~\cite{bai2025qwen3} & 8B
      & 7.9 & 43.90 & 153.16
      & 0.9 & 32.55 & 42.23 \\

    Qwen3.5-9B~\cite{bai2025qwen3} & 9B
      & 15.0 & 27.11 & 144.02
      & 2.6 & 19.51 & 26.40 \\

    InternVL3.5-HF~\cite{wang2025internvl3} & 8B
      & 13.8 & 23.49 & 134.61
      & 2.8 & 20.57 & 23.05 \\

    GLM-4.6V-Flash~\cite{hong2025glm} & 9B
      & 6.1 & 51.23 & 157.28
      & 0.9 & 92.81 & 102.38 \\

    DeepSeek-V4.1-Flash~\cite{xu2026deepseek} & 552B
      & \textbf{\textcolor{red}{22.2}} & \textbf{\textcolor{blue}{21.00}} & 139.75
      & \textbf{\textcolor{red}{8.3}} & \textbf{\textcolor{red}{7.89}} & \textbf{\textcolor{red}{11.04}} \\

    \addlinespace[3pt]
    \multicolumn{8}{c}{\cellcolor{green!8}\textit{Tool-augmented MLLMs}} \\

    Qwen3-VL + ZoomEye~\cite{shen2025zoomeye} & 8B
      & 7.4 & 35.36 & 147.19
      & 1.2 & 33.96 & 41.09 \\

    DeepEyes~\cite{zheng2026deepeyes} & 7B
      & 8.2 & 33.83 & 144.00
      & 1.3 & 26.86 & 31.89 \\

    Vision-OPD~\cite{yuan2026vision} & 9B
      & 12.4 & 31.19 & 144.43
      & 2.2 & 30.21 & 41.88 \\

    \addlinespace[3pt]
    \multicolumn{8}{c}{\cellcolor{orange!8}\textit{Ours}} \\

    \textbf{CountDraw} & 9B
      & \textbf{\textcolor{blue}{16.8}}
      & \textbf{\textcolor{red}{19.51}}
      & \textbf{\textcolor{red}{130.46}}
      & \textbf{\textcolor{blue}{5.6}}
      & \textbf{\textcolor{blue}{12.00}}
      & \textbf{\textcolor{blue}{16.41}} \\

    \bottomrule
  \end{tabular*}
  \endgroup
  \vspace{-3mm}
\end{table*}

\subsection{Ablation Study}
\label{sec:ablation}
Table~\ref{tab:countdraw-ablation} summarizes the reward ablations. Removing $R_\mathrm{box}$ reduces counting accuracy from 54.3\% to 52.4\%, while removing $R_\mathrm{zoom}$ leads to a larger drop to 50.3\%. The full method also achieves an MAE of 2.80, tied with the variant w/o $R_\mathrm{zoom}$. These results demonstrate that both rewards contribute to counting performance.

\subsection{Case Studies}
\label{sec:case}
\begin{figure}[t]
    \centering
    \includegraphics[width=\linewidth]{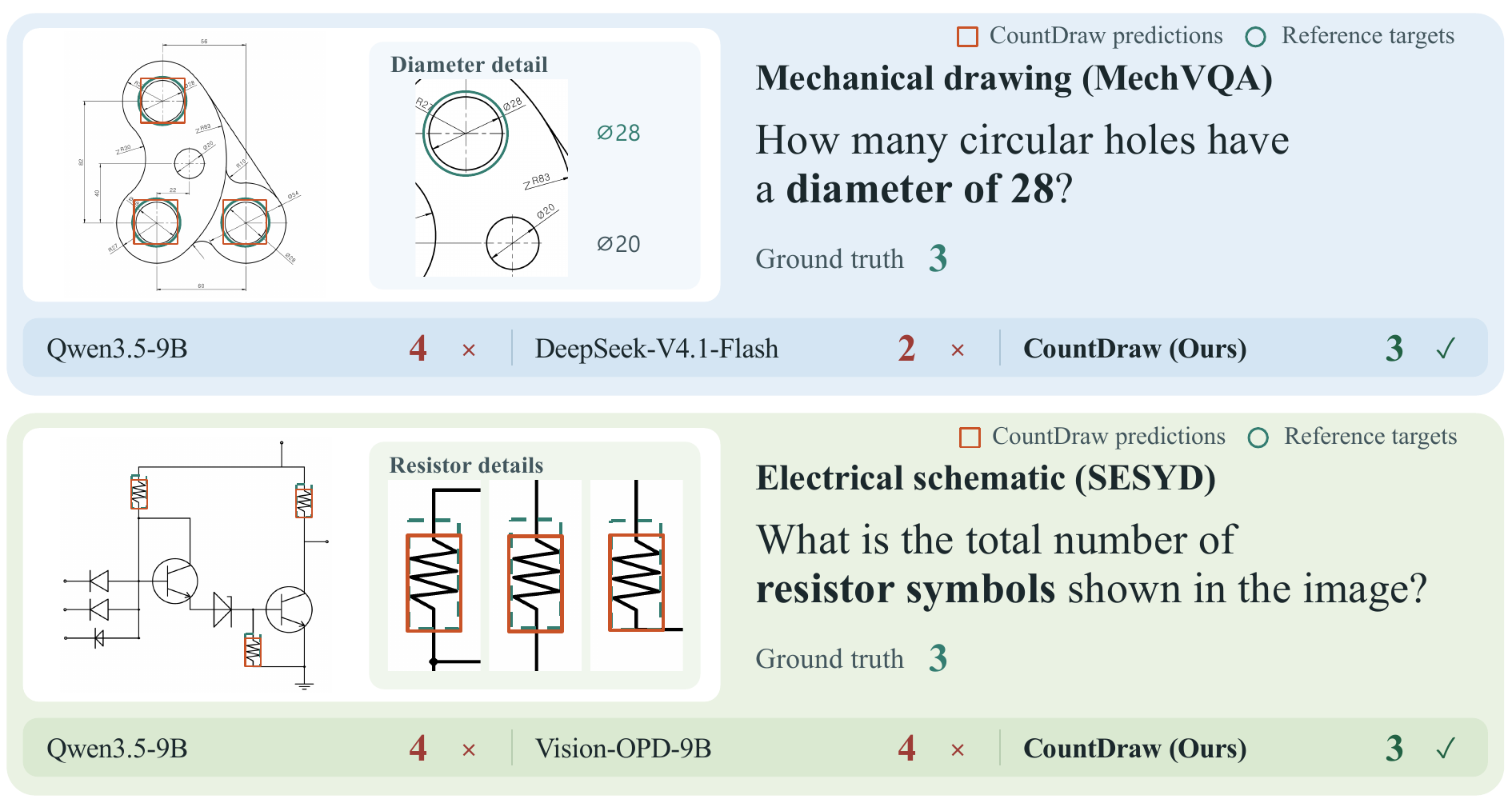}
    \caption{Case studies comparing CountDraw-9B with other baselines. CountDraw can provide the correct answer and give the correct localization (like top, bottom, bottom right).}
    \label{fig:case}
    \vspace{-5mm}
\end{figure}

Figure~\ref{fig:case} shows two examples from mechanical drawings and electrical schematics. In the mechanical drawing, Qwen3.5-9B counts more circular holes than truth, while DeepSeek-V4.1-Flash misses one. In the electrical schematic, both Qwen3.5-9B and Vision-OPD-9B incorrectly predict four resistors. CountDraw gives the correct count in both cases, with its predicted locations matching the reference targets. These examples further show that our method improves both counting accuracy and the ability to identify the correct objects in technical drawings.

\subsection{Evaluation on External Counting Benchmarks}
We further evaluate CountDraw on two established counting benchmarks, FSC-147~\cite{ranjan2021learning} and CARPK~\cite{hsieh2017drone}, to examine its generalization to other domains. FSC-147 covers diverse object categories and dense scenes, while CARPK focuses on counting vehicles in aerial images. Table~\ref{tab:countdraw-external} suggests that both benchmarks remain challenging. Even DeepSeek-V4.1-Flash achieves a low exact-count accuracy. 

Despite being trained on technical drawings, CountDraw-9B shows transfer to both datasets. Compared with its Qwen3.5-9B backbone, it reduces MAE by 28.0\% on FSC-147 and 38.5\% on CARPK, while also improving accuracy. CountDraw-9B also achieves the lowest MAE (19.51) and RMSE (130.46) among the evaluated open-source MLLMs on FSC-147. It is comparable to or better than the DeepSeek model with 552 billion parameters. These results show that the counting ability from technical drawings may generalize to natural images. 

\section{Conclusion}
\label{sec:conclusion}
In this paper, we study the challenging task of counting objects in technical drawings. We introduce CountDrawBench, a benchmark covering architectural plans, electrical schematics, mechanical drawings, and molecular structures. To improve counting performance, we further propose Code-Verified RL, which combines count verification, instance matching, and local zoom recounting to provide reliable training feedback. Extensive experiments show that current MLLMs still struggle with technical drawing counting, while our method achieves strong improvements over baselines. The gains also transfer to natural-image counting benchmarks. We hope CountDrawBench will support future research on visual counting and more reliable understanding of technical drawings.

\section*{LLM Usage Statement}
 We use LLMs to polish our language in our paper. We use codex to remove bugs in data creation and training, and create necessary scripts for execution. Codex is also used to check available GPUs and submit training tasks. We have carefully checked all things from LLMs. No data leakage, privacy issues and other issues happen. We do not use LLMs for other purposes.  We confirm that we obey all rules for LLM Usage.
{
    \small
    \bibliographystyle{ieeenat_fullname}
    \bibliography{main}
}

\clearpage
\appendix
\maketitlesupplementary

\section{Details of the Five Data Sources}
\label{app:data_source_detail}

We construct CountDrawBench from five data sources, covering architectural floor plans, electrical schematics, mechanical drawings, and molecular structures. Each source provides different types of technical drawings and reference annotations. We briefly introduce them below. 

\noindent \textbf{SESYD} SESYD~\cite{ranjan2021learning} is a synthetic document dataset developed for evaluating symbol recognition and spotting systems. It includes floor plans and electrical circuit diagrams containing multiple symbol categories with different shapes and arrangements. The dataset provides instance-level annotations, including symbol categories and spatial locations. We use its floor plan and circuit diagram collections to construct two subsets, SESYD FloorPlans and SESYD Circular, with counting questions derived from the annotated symbol instances. 

\noindent \textbf{FloorPlanCAD.} FloorPlanCAD~\cite{fan2021floorplancad} is a large-scale CAD drawing dataset designed for panoptic symbol spotting in architectural drawings. It contains floor plans with detailed annotations of architectural components, including their categories and bounding boxes. These drawings often contain repeated symbols distributed across different rooms and structural regions. We select annotated symbol categories and use their instance counts to construct the Floor-CAD subset.

\noindent \textbf{CubiCasa5K.} CubiCasa5K~\cite{kalervo2019cubicasa5k} is a floor plan dataset containing thousands of architectural layouts with annotations of rooms and building elements. Its annotations provide information about different room types and architectural components, including doors. We use the door-related annotations to construct counting questions involving different door categories, such as bedroom doors and kitchen doors. The resulting subset is referred to as Door. 

\noindent \textbf{ChemOCR.} We use molecular structure images and their corresponding reference structures from the ChemOCR~\cite{li2025chemvlm} data. These images represent molecules through chemical drawing conventions, where atoms and bonds are expressed using element labels, lines, and implicit structural representations. We parse the reference molecular structures using RDKit and extract counts for eight atom types, rings, aromatic rings, and double bonds. The resulting questions cover both explicitly labeled atoms and atoms implied by skeletal notation.

\noindent \textbf{MechVQA.} MechVQA~\cite{kou2026mechvqa} is a benchmark for multimodal understanding of mechanical engineering drawings. Its images contain engineering components, geometric features, dimensions, and multiple views of mechanical parts. We collect the counting questions provided by MechVQA and extend the annotations with additional object-counting questions.

\section{Counting Prompts}
\label{app:cd-counting-prompts}

Training examples with reference boxes use the localization
prompt. The rest use direct count. The zoom reward counts
each tile of a non-overlapping $2\times2$ grid.

\begingroup
\newcommand{\cdint}[1]{\texttt{\textbackslash boxed\{#1\}}}
\newcommand{\cdprompt}[2]{%
  \par\smallskip\noindent{\small\raggedright\textit{#1}\enspace #2\par}}
\cdprompt{Direct count.}{%
  Return only the final integer as \cdint{N}.}
\cdprompt{Count and localize.}{%
  Return \texttt{<bboxes>}\allowbreak
  \texttt{[[x1,y1,x2,y2],...]}\allowbreak
  \texttt{</bboxes>} followed by \cdint{N}.
  Use one tight box per counted instance and integer coordinates
  from 0 to 1000 relative to the full image. For zero instances,
  return an empty list. Give the answer directly.}
\cdprompt{Tile count.}{%
  Count \textit{[target]} in this image tile only. Do not count
  unrecognizable fragments. Return only the integer count as
  \cdint{N}. Do not output bounding boxes or explanations.}

\par\medskip\noindent\textbf{Prompt variants}\par\nobreak
\cdprompt{Point then count.}{%
  \textit{[Question]} Mark each target with one center point,
  then count the points. Use integer coordinates from 0 to 1000
  relative to the full image. Return
  \texttt{\{"points": [[x,y],...], "count": N\}}.
  For zero targets, use \texttt{[]} and 0.}
\cdprompt{One-shot count.}{%
  \textit{[Training image and question]} Answer: \cdint{K}.\\
  \textit{[Test image and question]} Return only the final integer
  as \cdint{N}.}
\endgroup

\section{More Statistics of Our Benchmark}
\label{app:statistic}

Table~\ref{tab:benchmark-statistics} further shows some statistics of our CountDrawBench. We include metrics about both the textual query-answer pairs and the images. 

\begin{table}[t]
  \centering
  \resizebox{\linewidth}{!}{
  \begin{tabular}{lr}
    \hline
    Statistic & Value \\
    \hline
    Questions & 1,574 \\
    Distinct images & 707 \\
    Mean queries per image & 2.23 \\
    Mean resolution (px) & $2155.4\times1629.5$ \\
    Min. resolution (px) & $363\times162$ \\
    Max. resolution (px) & $6623\times4678$ \\
    Mean image area (MP) & 4.89 \\
    Mean query length (tokens) & 30.03 \\
    Min. query length (tokens) & 15 \\
    Max. query length (tokens) & 66 \\
    Mean GT count & 4.66 \\
    GT count, min--max & 0--76 \\
    Questions with GT count $>5$ & 522 \\
    Questions with GT boxes & 722  \\
    GT box annotations & 3,337 \\
    Mean boxes per bbox query & 4.62 \\
    \hline
  \end{tabular}
  }
  \caption{Statistics of CountDrawBench. Image sizes use distinct original images; minimum and maximum are selected by pixel area. Query lengths use the Qwen3.5-9B tokenizer on raw user text without special tokens. Box statistics use only bbox-annotated questions.}
  \label{tab:benchmark-statistics}
\end{table}

\section{Benchmark Difficulty}
\label{app:benchmark_difficulty}
\begin{figure*}[t]
  \centering
  \begin{minipage}[t]{0.48\linewidth}
    \centering
    \includegraphics[width=\linewidth]{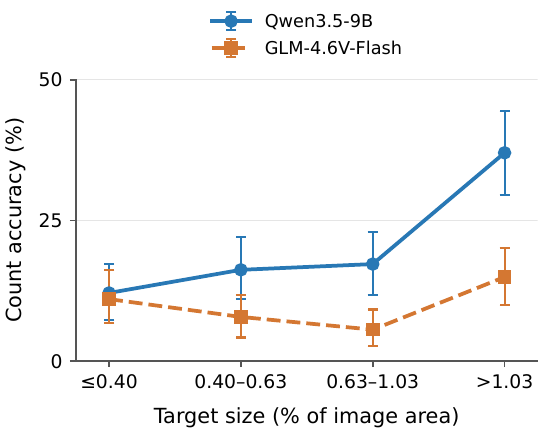}
  \end{minipage}\hfill
  \begin{minipage}[t]{0.48\linewidth}
    \centering
    \includegraphics[width=\linewidth]{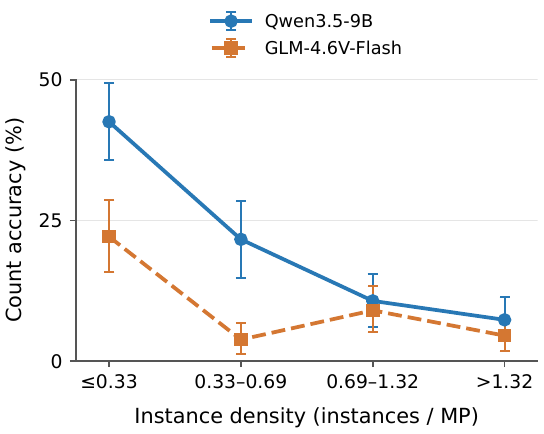}
  \end{minipage}
  \caption{Exact count accuracy of Qwen3.5-9B and GLM-4.6V-Flash on all
    722 bbox-annotated questions, grouped by target size and instance density.
    Target size is the mean instance-box area relative to image area;
    density is the number of queried instances per original-image megapixel.
    Bins use GT quartiles. Error bars show 95\% image-cluster bootstrap
    intervals; invalid count outputs are scored as incorrect.}
  \label{fig:gt-size-density}
\end{figure*}

Figure~\ref{fig:gt-size-density} further demonstrates the difficulties of our benchmark. We can discover that, as the target size dicreases and instance density increases, performances of both Qwen3.5-9B and GLM-4.6V-Flash drop very fast to a low score next to zero. Other MLLMs see a similar trend. Our benchmark therefore reveals a common weakness across (especially open-source) MLLMs. They may understand the technical drawings not so well and do the wrong counting. 

\section{Reward Ablation and RL-based Baselines}
\label{app:cd-perception-rewards}

\noindent \textbf{Further ablation on the reward.} We train Qwen3.5-9B on \textsc{CountDraw-7K} for 125 GRPO
updates with 56 questions per update and eight rollouts per
question. The count-only variant uses $R_{\mathrm{cnt}}$ alone.
The test split (1,574 questions) and metrics follow
Table~\ref{tab:countdraw-main}. The ablation results are in Table~\ref{tab:cd-supp-count-ablation}. Count-only training causes a 5.2 point degrade. Together with Table~\ref{tab:countdraw-ablation} in the main text, we can see that full Code-Verified RL training gains the most. 
\begin{table}[t]
  \caption{Count-reward ablation on CountDrawBench.}
  \label{tab:cd-supp-count-ablation}
  \centering
  \renewcommand{\arraystretch}{1.12}
  \begin{tabular}{@{\extracolsep{\fill}}lcc@{}}
    \toprule
    Method & Overall (\%) $\uparrow$ & MAE $\downarrow$ \\
    \midrule
    CountDraw & 54.3 & 2.80 \\
    Count only & 49.1 & 3.09 \\
    \bottomrule
  \end{tabular}
\end{table}

\noindent \textbf{Perception RL baselines.}
We further choose some RL-based baselines to support the effectiveness of CountDraw. Visual-RFT~\cite{liu2025visual} baseline awards one point each for an exact count and valid format. Perception-R1~\cite{yu2026perception} matches centers Euclidean distance and
weights the score by the ratio of the smaller to the larger count.
VisionReasoner~\cite{liu2026visionreasoner} tests box IoU ($>0.5$), box $L_1$ error
($<10$ pixels), and center $L_1$ error ($<30$ pixels) for each
matched pair. It divides the number of satisfied criteria by the
larger count and adds the format reward. 

Detailed results are in Table~\ref{tab:cd-supp-rl-dev}. We discover that CountDraw-9B achieves the \textbf{best accuracy} compared with all RL-based methods, and this is the most important metric. It also suggests that all RL-based methods achieve lower MAE on our benchmark, and CountDraw achieves the second best. Reinforcement Learning approaches could be promising in solving these fine-grained counting tasks.

\begin{table}[t]
  \caption{Perception RL results on CountDrawBench.}
  \label{tab:cd-supp-rl-dev}
  \centering
  \setlength{\tabcolsep}{2pt}
  \renewcommand{\arraystretch}{1.12}
  \begin{tabular}{lccc}
    \toprule
    Method & Number & Overall (\%) $\uparrow$ & MAE $\downarrow$ \\
    \midrule
    Visual-RFT     & 1,574 & 45.7 & 2.95 \\
    Perception-R1  & 1,574 & 50.0 & 2.61 \\
    VisionReasoner & 1,574 & 49.1 & 2.84 \\
    \midrule
    CountDraw & 1,574 & \textbf{54.3} & 2.80 \\
    \bottomrule
  \end{tabular}
  \vspace{-5mm}
\end{table}

\begin{table*}[t]
  \caption{Localization on questions with correct counts. mIoU is the mean matched IoU
    over all questions with correct ``\textit{final}'' count answers. Low IoU denotes a matched IoU below $0.5$.}
  \label{tab:cd-supp-correct-count-localization}
  \centering\footnotesize
  \renewcommand{\arraystretch}{1.22}
  \newcommand{\cdlow}[2]{#1\hspace{.4em}(#2\%)}
  \begin{tabular*}{\linewidth}{@{\extracolsep{\fill}}lrrr@{}}
    \toprule
    Model & \shortstack[r]{Exact\\counts~$\uparrow$}
      & \shortstack[r]{mIoU~$\uparrow$\\($\times 100$)}
      & \shortstack[r]{Correct counts\\but low IoU~$\downarrow$} \\
    \midrule
    \textbf{CountDraw} & 283 & 65.25 & \cdlow{70/283}{24.7} \\
Gemini-3.8-Flash~\cite{gemini3_8} & 296 & 49.71 & \cdlow{135/296}{45.6} \\
DeepSeek-V4.1-Flash~\cite{xu2026deepseek} & 217 & 15.88 & \cdlow{178/217}{82.0} \\
Qwen3-VL-8B-Instruct~\cite{bai2025qwen3} & 140 & 54.39 & \cdlow{58/140}{41.4} \\
Qwen3.5-9B~\cite{bai2025qwen3} & 203 & 23.73 & \cdlow{154/203}{75.9} \\
InternVL3.5-8B-HF~\cite{wang2025internvl3} & 117 & 23.83 & \cdlow{87/117}{74.4} \\
GLM-4.6V-Flash~\cite{hong2025glm} & 164 & 0.00 & \cdlow{164/164}{100.0} \\
Qwen3-VL + ZoomEye~\cite{shen2025zoomeye} & 165 & 22.59 & \cdlow{132/165}{80.0} \\
DeepEyes-7B~\cite{zheng2026deepeyes} & 160 & 0.00 & \cdlow{160/160}{100.0} \\
Vision-OPD-9B~\cite{yuan2026vision} & 181 & 22.10 & \cdlow{144/181}{79.6} \\
    \bottomrule
  \end{tabular*}
\end{table*}

\section{Localization Given a Correct Count}
\label{app:cd-correct-count-localization}
Table~\ref{tab:cd-supp-correct-count-localization} further examines whether models identify the correct instances when their predicted counts are correct. We observe that accurate counting does not always indicate accurate localization. For example, CountDraw and Gemini-3.8-Flash correctly answer a similar number of questions, but their localization results differ substantially. CountDraw achieves the highest matched IoU of 65.25, with only 24.7\% of its correct-count predictions having low IoU, compared with 45.6\% for Gemini. These results show that Code-Verified RL helps the model identify the correct target instances, even when counting accuracy alone shows little difference.

\section{Grounded Counting Examples}
\label{app:cd-grounding-examples}
Figure~\ref{fig:cd-supp-grounded-examples} and Figure~\ref{fig:cd-supp-grounded-examples-more} show some cases with ground truth. On these cases, our CountDraw-9B can provide the correct counting answer that other MLLMs fail to. 

\begingroup
\makeatletter
\newcommand{\cdcaseshade}[1]{%
  \rlap{\color{#1}\rule[-\dp\@arstrutbox]{\linewidth}%
    {\dimexpr\ht\@arstrutbox+\dp\@arstrutbox\relax}}}
\makeatother
\newcommand{\cdcasetable}[1]{%
  \par\smallskip\noindent
  {\small
   \setlength{\tabcolsep}{1.5pt}%
   \renewcommand{\arraystretch}{1.30}%
   \begin{tabular*}{\linewidth}{@{\extracolsep{\fill}}lrrrl@{\hspace{3pt}}}
     \toprule
     \hspace{3pt}\rlap{Model}\phantom{Qwen3-VL + ZoomEye}
       & Final count & Matched & Extra & \rlap{Outcome}\phantom{Box error} \\
     \midrule
     #1%
     \bottomrule
   \end{tabular*}\par}}
\newcommand{\cdcaserow}[6]{%
  \cdcaseshade{#1}\hspace{3pt}#2 & #3 & #4 & #5 & #6\\}
\newlength{\cdartheight}
\setlength{\cdartheight}{143bp}
\newcommand{\cdcasepanel}[6][]{%
  \begin{minipage}[t]{\columnwidth}%
    \small\raggedright
    \noindent#1\textbf{#2}\strut\par\smallskip
    \noindent\vbox to \cdartheight{%
      \vfil\hbox to \linewidth{\hfil#3\hfil}\vfil}\par\smallskip
    \noindent\parbox[t][2\baselineskip][t]{\linewidth}{%
      \raggedright\strut\textit{Question.} #4\strut}\par
    \noindent\textit{Reference count.} #5\strut\par
    \cdcasetable{#6}%
  \end{minipage}}
\newcommand{\cdcaselegend}{%
  \par\smallskip\noindent
  \makebox[\linewidth][r]{\makebox[0pt][r]{\makebox[\columnwidth][c]{%
    \footnotesize\setlength{\unitlength}{1bp}\linethickness{.85bp}%
    \textcolor[RGB]{35,95,159}{\begin{picture}(15,7)%
      \put(.5,.5){\framebox(14,6){}}\end{picture}}%
    \hspace{.45em}CountDraw prediction\hspace{1.8em}%
    \textcolor[RGB]{36,113,63}{\begin{picture}(15,7)%
      \put(.5,.5){\dashbox{2}(14,6){}}\end{picture}}%
    \hspace{.45em}Reference annotation}}}\par}

\noindent
\cdcasepanel[\refstepcounter{figure}\label{fig:cd-supp-grounded-examples}]{(a) Tables in a floor plan}
  {\includegraphics[trim={0bp 292bp 0bp 27bp},clip,width=\linewidth,height=\cdartheight,keepaspectratio]
     {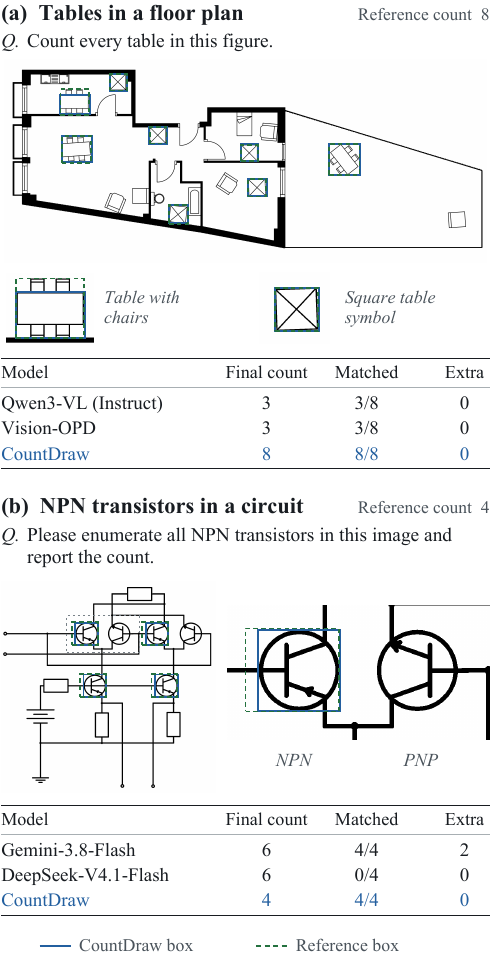}}
  {Count every table in this figure.}{8}{%
    \cdcaserow{red!7}{Qwen3-VL-8B-Instruct}{3}{3/8}{0}{Incorrect}
    \cdcaserow{red!7}{Vision-OPD-9B}{3}{3/8}{0}{Incorrect}
    \cdcaserow{green!7}{CountDraw}{8}{8/8}{0}{Correct}}%
\par
\newpage
\vspace*{10.3pt}
\noindent
\cdcasepanel{(b) NPN transistors in a circuit}
  {\includegraphics[trim={0bp 77bp 0bp 277bp},clip,width=\linewidth,height=\cdartheight,keepaspectratio]
     {Figures/countdraw-grounding-details.pdf}}
  {Please enumerate all NPN transistors in this image and report the count.}{4}{%
    \cdcaserow{red!7}{Gemini-3.8-Flash}{6}{4/4}{2}{Incorrect}
    \cdcaserow{red!7}{DeepSeek-V4.1-Flash}{6}{0/4}{0}{Incorrect}
    \cdcaserow{green!7}{CountDraw}{4}{4/4}{0}{Correct}}%
\cdcaselegend

\noindent
\cdcasepanel[\refstepcounter{figure}\label{fig:cd-supp-grounded-examples-more}]{(c) Views of a plunger pump}
  {\includegraphics[page=1,width=\linewidth,height=\cdartheight,keepaspectratio]
     {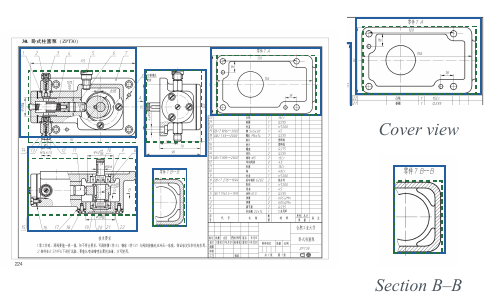}}
  {How many views does this drawing contain?}{5}{%
    \cdcaserow{red!7}{Vision-OPD-9B}{4}{0/5}{0}{Incorrect}
    \cdcaserow{red!7}{InternVL3.5-8B-HF}{4}{0/5}{1}{Incorrect}
    \cdcaserow{green!7}{CountDraw}{5}{5/5}{0}{Correct}}%
\hfill
\cdcasepanel{(d) Aromatic rings in a molecule}
  {\includegraphics[page=2,width=\linewidth,height=\cdartheight,keepaspectratio]
     {Figures/countdraw-cross-domain-cases.pdf}}
  {Please enumerate all aromatic rings in this molecule and report the count.}{2}{%
    \cdcaserow{red!7}{InternVL3.5-8B-HF}{1}{0/2}{1}{Incorrect}
    \cdcaserow{red!7}{Qwen3-VL + ZoomEye}{2}{0/2}{2}{Box error}
    \cdcaserow{green!7}{CountDraw}{2}{2/2}{0}{Correct}}%
\par
\vspace{14pt}
\noindent
\cdcasepanel{(e) Wardrobes in a CAD drawing}
  {\includegraphics[page=3,width=\linewidth,height=\cdartheight,keepaspectratio]
     {Figures/countdraw-cross-domain-cases.pdf}}
  {Please enumerate all wardrobes in this image and report the count.}{3}{%
    \cdcaserow{red!7}{Gemini-3.8-Flash}{0}{0/3}{0}{Incorrect}
    \cdcaserow{red!7}{DeepSeek-V4.1-Flash}{5}{3/3}{2}{Incorrect}
    \cdcaserow{green!7}{CountDraw}{3}{3/3}{0}{Correct}}%
\hfill
\cdcasepanel{(f) Diodes in a circuit}
  {\includegraphics[page=4,width=\linewidth,height=\cdartheight,keepaspectratio]
     {Figures/countdraw-cross-domain-cases.pdf}}
  {Report the exact count of diode symbols in this image.}{5}{%
    \cdcaserow{red!7}{Qwen3-VL-8B-Instruct}{7}{3/5}{4}{Incorrect}
    \cdcaserow{red!7}{InternVL3.5-8B-HF}{3}{0/5}{3}{Incorrect}
    \cdcaserow{green!7}{CountDraw}{5}{5/5}{0}{Correct}}%
\par
\cdcaselegend
\vspace{12pt}
\endgroup

\section{Annotation Agreement}
\label{app:annotation_agree}

Our human-annotated subset of Mech in CountDrawBench include 502 data samples. 1242 Mech samples are in the training set. Our four annotators are all graduated students, and this annotation (only the numbers of instances are needed) is easy for them. They do all the annotations individually. The full-agreement rate of this annotation is 90.7\% and Krippendorff's $\alpha$~\cite{krippendorff2011computing} (a metric measuring multi-people agreement) is 0.895. All metrics suggest strong agreement of the annotation. 

\section{Ethical Consideration}
\label{app:ethical}
CountDrawBench is constructed from data sources of technical drawings, including floor plans, circuit diagrams, mechanical drawings, and molecular structures. The benchmark focuses on object counting and does not involve personal or sensitive information. We use human annotators to verify counting labels in mechanical drawings. While Code-Verified RL improves counting and localization performance, the model may still make mistakes. Its predictions should be checked by human experts when used in engineering or other applications where counting errors may have serious consequences. 
\end{document}